\documentclass[lettersize,journal]{IEEEtran}
\usepackage{amsmath,amsfonts}
\usepackage{algorithmic}
\usepackage{algorithm}
\usepackage{array}
\usepackage[caption=false,font=normalsize,labelfont=sf,textfont=sf]{subfig}
\usepackage{textcomp}
\usepackage{stfloats}
\usepackage{url}
\usepackage{verbatim}
\usepackage{graphicx}
\usepackage{cite}

\usepackage{multirow}
\usepackage{amsthm}
\usepackage[table]{xcolor}
\usepackage{booktabs}

\newcommand{\methodName}{ARDepth}
\begin{document}

\title{ARDepth: Auto-regressive Monocular Depth Estimation with Progressive Visual Conditioning}

\author{
Zijie~Wang,
Wei~Zhang,
Weiming~Zhang,
Xiao~Tan,
Weikai~Chen,
Xiaoxu~Li,
Guanbin~Li

\IEEEcompsocitemizethanks{\IEEEcompsocthanksitem Zijie Wang and Guanbin Li are with the School of Computer Science and Engineering, Sun Yat-sen University, Guangzhou, 510006, China. They are also with Shenzhen Loop Area Institute, Shenzhen, 518038, China. Guanbin Li is also with Guangdong Key Laboratory of Big Data Analysis and Processing, Guangzhou 510006, China.
\IEEEcompsocthanksitem Wei Zhang, Weiming Zhang and Xiao Tan are with Baidu Inc., Shenzhen, 518066, China. 
\IEEEcompsocthanksitem WeiKai Chen is with LightSpeed Studios, Tencent America.
\IEEEcompsocthanksitem Xiaoxu Li is with the School of Computer Science and Artificial Intelligence, Lanzhou University of Technology, Lanzhou 730050, China.
\IEEEcompsocthanksitem  Corresponding Author is Guanbin Li~(liguanbin@mail.sysu.edu.cn).
}
}




\maketitle

\begin{abstract}
Diffusion models have recently become the dominant paradigm for monocular depth estimation (MDE). However, they implicitly assume that depth can be recovered as a globally smooth field through iterative denoising, which does not explicitly reflect the piecewise and scale-dependent organization of scene geometry. In practice, geometric structure emerges progressively across spatial scales, where coarse layout, surfaces, and boundaries are constructed in a hierarchical manner.
Motivated by this observation, we introduce \methodName{}, which formulates depth estimation as structured auto-regressive generation. Instead of recovering depth through global refinement, \methodName{} progressively constructs depth representations as spatial resolution increases.
To support this generative process, we introduce Scale-Progressive Conditioning (SPC) to inject multi-scale visual features at each generation stage, and Semantic-Aware Guidance (SAG) to provide scene-level semantic priors that enhance global structural consistency.
Together, these designs enable the model to capture fine-grained local details while maintaining coherent global geometry.
Empirical results demonstrate that our approach achieves strong performance and produces structurally consistent depth predictions across scales, validating auto-regressive generation as a promising alternative paradigm for geometric modeling.
\end{abstract}


\section{Introduction}
\label{sec:intro}

\IEEEPARstart{I}n recent years, diffusion-based models~\cite{Marigold,he2024lotus,depthmaster} have defined the dominant paradigm for monocular depth estimation (MDE). These approaches achieve strong zero-shot generalization and consistent empirical gains by modeling depth as a noisy field that can be progressively refined through iterative denoising. Despite their success, diffusion models are based on the assumption that depth can be treated as a globally coherent random field, whose structure can be gradually uncovered through uniform denoising. 
However, geometric scenes are inherently \textit{piecewise and discontinuous}, with sharp transitions at occlusion boundaries, surface intersections and structural changes. While uniform denoising provides a powerful optimization mechanism, it does not explicitly reflect the hierarchical organization of geometric complexity across spatial scales.

\begin{figure*}[t]
    \centering
    \includegraphics[width=\linewidth]{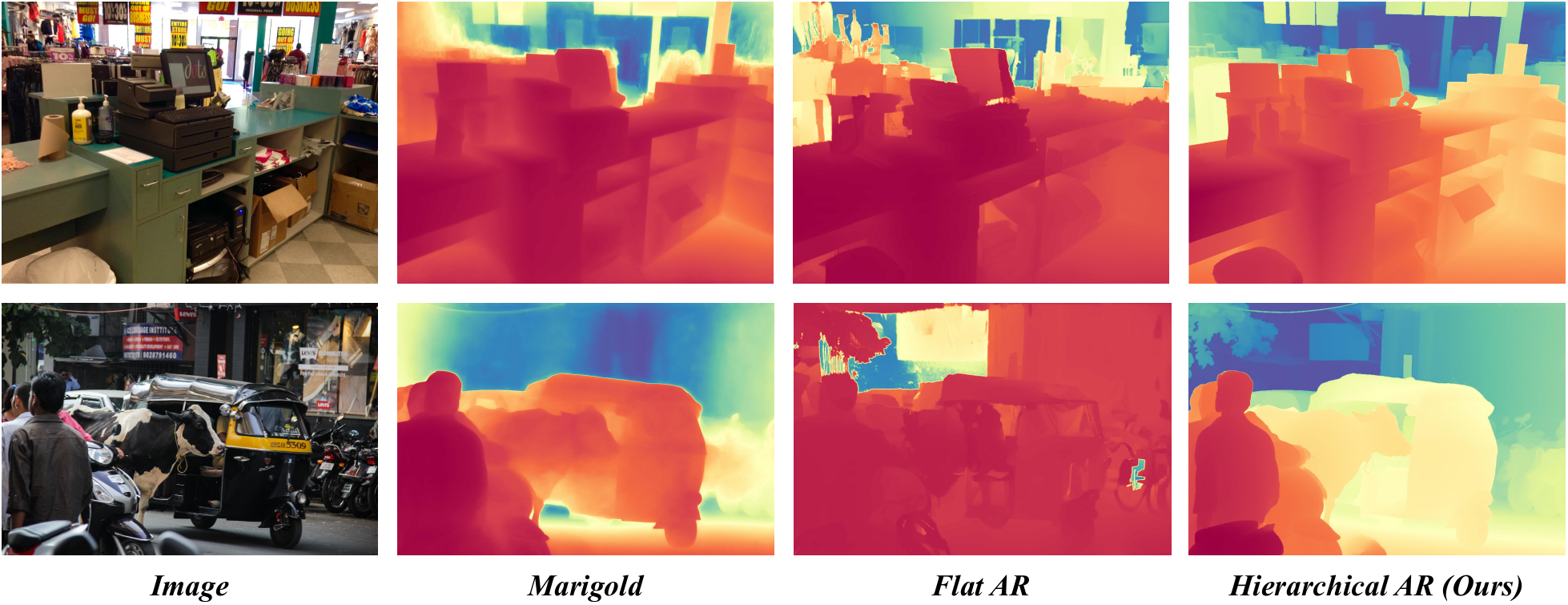}
    \caption{\textbf{Motivation.} Given an input RGB image, we compare a typical diffusion-based model (Marigold~\cite{Marigold}), a flat auto-regressive (AR) baseline, and our hierarchical AR framework. In these challenging examples, diffusion-based depth prediction shows typical difficulties in preserving sharp discontinuities and thin structures, which are known hard cases for its iterative denoising paradigm. The flat AR baseline captures the general spatial layout and object placement but struggles to resolve fine-grained depth variations without a scale hierarchy. Our hierarchical coarse-to-fine AR formulation progressively refines geometry across resolutions, yielding sharper boundaries, improved local detail and more coherent depth in these difficult scenarios.}
    \label{fig:motivation}
\end{figure*}

From a structural perspective, the examples in Fig.~\ref{fig:motivation} illustrate a common challenge of diffusion-based depth estimation: it remains difficult to preserve sharp discontinuities and thin structures, which are known hard cases for iterative denoising.
This observation suggests that depth may not always be best modeled as a globally refined random field. 
Instead, geometric structure often emerges through the progressive composition of surfaces, boundaries and occlusions.
Auto-regressive (AR) modeling provides a constructive generative mechanism that sequentially builds representations conditioned on previously generated structure. 
Such a mechanism offers a natural alternative for modeling geometric signals, as it enables spatial structure to be incrementally composed rather than uniformly refined. 

However, applying AR modeling to dense geometric prediction is non-trivial, as it requires effective mechanisms to model long-range spatial dependencies while preserving fine-grained structural details. Empirically, we observe that a single-scale (flat) AR formulation still struggles to allocate sufficient capacity to near-field details, even when the overall layout is captured correctly, as illustrated in Fig.~\ref{fig:motivation}. 
This limitation suggests that effective AR-based depth modeling must explicitly account for the scale-dependent nature of geometric complexity.

We instantiate this perspective in \methodName{}, a hierarchical AR framework for monocular depth estimation.
\methodName{} unfolds depth auto-regressively from coarse to fine resolution, such that representational capacity increases in tandem with geometric complexity.
To make this generative schedule effective in practice, \methodName{} integrates two complementary mechanisms.
The Scale-Progressive Conditioning (SPC) module constructs multi-level visual signals from the input image and injects stage-specific evidence at each auto-regression step, ensuring that coarse stages receive global layout cues while high-resolution stages are guided by more localized visual information.
The Semantic-Aware Guidance (SAG) module further introduces lightweight scene-level semantics, which help stabilize global structure and reduce ambiguity in textureless or occluded regions.
Together, these components encourage geometry to be progressively refined in a manner consistent with its scale-dependent organization.
Empirically, as illustrated in Fig.~\ref{fig:motivation}, \methodName{} produces sharper boundaries and more faithful near-field geometry compared with representative diffusion-based models, while retaining strong zero-shot transfer performance.

In summary, our contributions are as follows:
\begin{itemize}
    \item \textit{Structured AR formulation for MDE}. We formulate monocular depth estimation as scale-progressive auto-regressive generation, where depth is constructed from coarse layout to fine geometric details rather than recovered through uniform denoising.
    \item \textit{Stage-aligned conditioning}. We introduce Scale-Progressive Conditioning and Semantic-Aware Guidance to align visual and semantic cues with the hierarchical generation process, enabling local detail recovery while maintaining global scene coherence.
    \item \textit{Comprehensive validation}. Extensive zero-shot, relative-depth, in-domain, scaling, and ablation studies show that structured AR generation is a competitive and scalable alternative for dense geometric prediction.
\end{itemize}

\section{Related Work}
\label{sec:related_work}

\subsection{Monocular Depth Estimation}

Monocular depth estimation (MDE) aims to recover dense scene geometry from a single RGB image. 
As a fundamental geometric perception task, MDE is broadly useful for downstream scene understanding~\cite{philion2020lift,li2023bevdepth,dan2026optimvmap,wang2025lanediffusion}.
Early learning-based methods are usually trained and evaluated within specific domains~\cite{xu2018pad,song2023ec,liu2024plane2depth}, achieving strong performance on standard benchmarks but showing limited generalization to unseen environments. 
To improve cross-domain robustness, later works explore relative or affine-invariant depth estimation, where predictions are evaluated up to unknown scale and shift. 
MiDaS~\cite{MiDas} and related multi-dataset training strategies show that mixing heterogeneous data sources can substantially improve zero-shot generalization.

Recent progress has increasingly pushed MDE toward large-scale foundation models. 
The Depth Anything family~\cite{DA,DA2} demonstrates that large-scale labeled and pseudo-labeled data can provide a powerful path toward robust zero-shot depth estimation. 
Metric-oriented methods further extend this direction. 
UniDepth~\cite{piccinelli2024unidepth} targets universal monocular metric depth estimation from a single image without requiring additional scene or camera information. 
Metric3D v2~\cite{hu2024metric3dv2} studies zero-shot metric depth and surface normal estimation by addressing camera-model ambiguity and leveraging large-scale geometric supervision. 
Depth Pro~\cite{bochkovskii2025depthpro} emphasizes sharp metric depth prediction with high-resolution details and fast inference. 
These works highlight the importance of data scale, geometric normalization, and foundation-model pretraining for robust depth estimation.

Despite their strong performance, data-scaling methods often depend on large training corpora, pseudo-labeling pipelines, or metric-depth-specific designs. 
A complementary research direction explores whether pretrained generative models can provide transferable priors for dense geometric prediction with more moderate depth-specific supervision. 
Our work follows this broader motivation, but studies a different generative formulation: instead of treating depth prediction mainly as discriminative regression or diffusion-based refinement, \methodName{} formulates MDE as a scale-progressive auto-regressive generation process.

\subsection{Diffusion-based Generative Priors for Depth Estimation}

Diffusion-based models have recently shown strong potential for monocular depth estimation. 
Marigold~\cite{Marigold} adapts a pretrained Stable Diffusion model to affine-invariant depth prediction, showing that image-generation priors can be repurposed for dense geometric perception. 
This line has inspired several follow-up works. 
Lotus~\cite{he2024lotus} analyzes diffusion formulations for dense prediction and simplifies the process toward efficient annotation prediction. 
DepthFM~\cite{depthfm} studies efficient flow-matching-based depth generation. 
DepthMaster~\cite{depthmaster} adapts diffusion features for discriminative depth estimation and enhances both global structure and fine details. 
PriorDiffusion~\cite{zeng2024priordiffusion}, GeoWizard~\cite{fu2024geowizard}, and GenPercept~\cite{GenPercept} further explore diffusion priors for dense perception through different conditioning, adaptation, and inference strategies. 
Beyond pure MDE, Marigold-DC~\cite{viola2025marigolddc} extends Marigold-style diffusion priors to zero-shot depth completion, indicating that image-conditioned diffusion priors can also guide sparse-to-dense geometric reconstruction.

These methods validate the value of pretrained generative priors for dense prediction. 
At the same time, most of them still rely on a denoising or refinement view of depth generation, even when the process is shortened or reformulated for efficiency. 
Such a formulation is effective for producing coherent dense maps, but it does not explicitly describe how geometric structures are constructed across spatial scales. 
Depth maps are piecewise and discontinuous, with sharp transitions at occlusion boundaries, thin structures, surface intersections, and object contours. 
This motivates us to explore a complementary generative perspective, where depth is progressively constructed from coarse layout to fine geometric details.

\subsection{Visual Auto-regressive Modeling}

Auto-regressive modeling has achieved remarkable success in language modeling~\cite{qwen2.5-VL} and has recently been extended to visual generation~\cite{van2017neural,mao2025VAREdit,wang2026world}. 
Early visual AR models often discretize images into tokens through vector quantization~\cite{van2017neural,esser2021taming,chang2022maskgit}, followed by sequential token prediction with Transformer architectures. 
Later works improve tokenization and generation quality through lookup-free quantization and more scalable visual token modeling~\cite{yu2023language,zhao2024image}. 
VAR~\cite{tian2024visual} reformulates image generation as next-scale prediction rather than raster-scan next-token prediction, providing a coarse-to-fine visual AR paradigm. 
Infinity~\cite{han2025infinity} further studies bitwise visual auto-regressive modeling for high-resolution image synthesis. 
Recent editing models such as Editar~\cite{mu2025editar} and VAREdit~\cite{mao2025VAREdit} adapt visual AR modeling to image editing, where conditioning on source content is important for preserving structure while generating target outputs.

Although these works show the scalability of visual AR modeling, most of them focus on image synthesis or editing. 
Dense geometric prediction imposes different requirements. 
MDE requires pixel-aligned outputs, accurate object boundaries, coherent surface structures, and globally consistent depth ordering. 
Directly applying generic visual AR modeling to depth prediction can be insufficient because the generated sequence must preserve geometric correspondence rather than only semantic plausibility.

Recent works have begun to explore auto-regressive depth estimation. 
DepthART~\cite{gabdullin2024depthart} adapts Visual Auto-Regressive Transformer to MDE and introduces a Depth Autoregressive Refinement Task to mitigate the gap between teacher-forced training and auto-regressive inference. 
DAR~\cite{wang2025dar} formulates MDE as a scalable auto-regressive prediction problem by modeling depth maps across resolutions and depth granularities. 
These works confirm that AR modeling is a promising direction for depth estimation. 
Different from directly adapting generic AR objectives, \methodName{} focuses on task-aware hierarchical generation, where depth residuals are progressively constructed across scales and guided by stage-aligned visual and semantic conditions.

\subsection{Conditioning for Dense Prediction}

Effective conditioning is crucial for dense prediction, where models must preserve both global scene layout and local spatial details. 
A common strategy is to exploit multi-scale visual representations. 
Feature Pyramid Networks~\cite{lin2017feature} combine semantically strong coarse features with spatially detailed fine features and have become a standard design for dense visual tasks. 
In monocular depth estimation, BTS~\cite{lee2019big} introduces multi-scale local planar guidance during decoding, while DPT~\cite{DPT} assembles transformer features into dense multi-resolution predictions. 
These works suggest that depth estimation should not rely on a single-level representation, since scene layout, object structures, and local depth discontinuities are often captured at different spatial scales.

Recent generative and foundation-model-based approaches further highlight the importance of conditioning. 
Diffusion-based depth models condition generation on image features or adapted latent representations to transfer pretrained generative priors to dense geometry~\cite{Marigold,he2024lotus,depthfm,depthmaster}. 
Visual auto-regressive models also require conditioning signals to guide sequential prediction~\cite{tian2024visual,han2025infinity,mao2025VAREdit}. 
However, dense geometry imposes stricter spatial requirements than image-level generation or editing. 
Global conditioning alone may weaken local boundaries and thin structures, while purely local cues may fail to preserve coherent scene layout. 
This motivates conditioning designs that are both spatially aligned and aware of the current generation scale.

Language and vision-language priors have also been explored for depth estimation. 
DepthCLIP~\cite{zhang2022can} transfers CLIP's semantic responses to coarse depth prediction through depth-bin prompts, and subsequent prompt-learning variants further study how language prompts can adapt CLIP to MDE~\cite{auty2023learning}. 
WorDepth~\cite{zeng2024wordepth} explicitly models language as a prior for monocular depth by exploiting object-level semantic and scale cues from captions. 
Vision-Language Embodiment~\cite{zhang2025visionlanguageembodiment} incorporates text descriptions as scene-understanding priors together with geometric cues. 
Relatedly, VPD~\cite{zhao2023unleashing} shows that pretrained text-to-image diffusion models contain transferable representations for dense visual perception, including depth estimation.
These works indicate that language can provide useful semantic context, but also suggest that fine-grained textual depth reasoning may be unreliable or ambiguous. 
Therefore, in \methodName{}, language is used as compact scene-level guidance rather than direct pixel-level depth supervision, while scale-aligned visual cues remain responsible for local geometric reconstruction.

\begin{figure*}[!t]
    \centering
    \includegraphics[width=\linewidth]{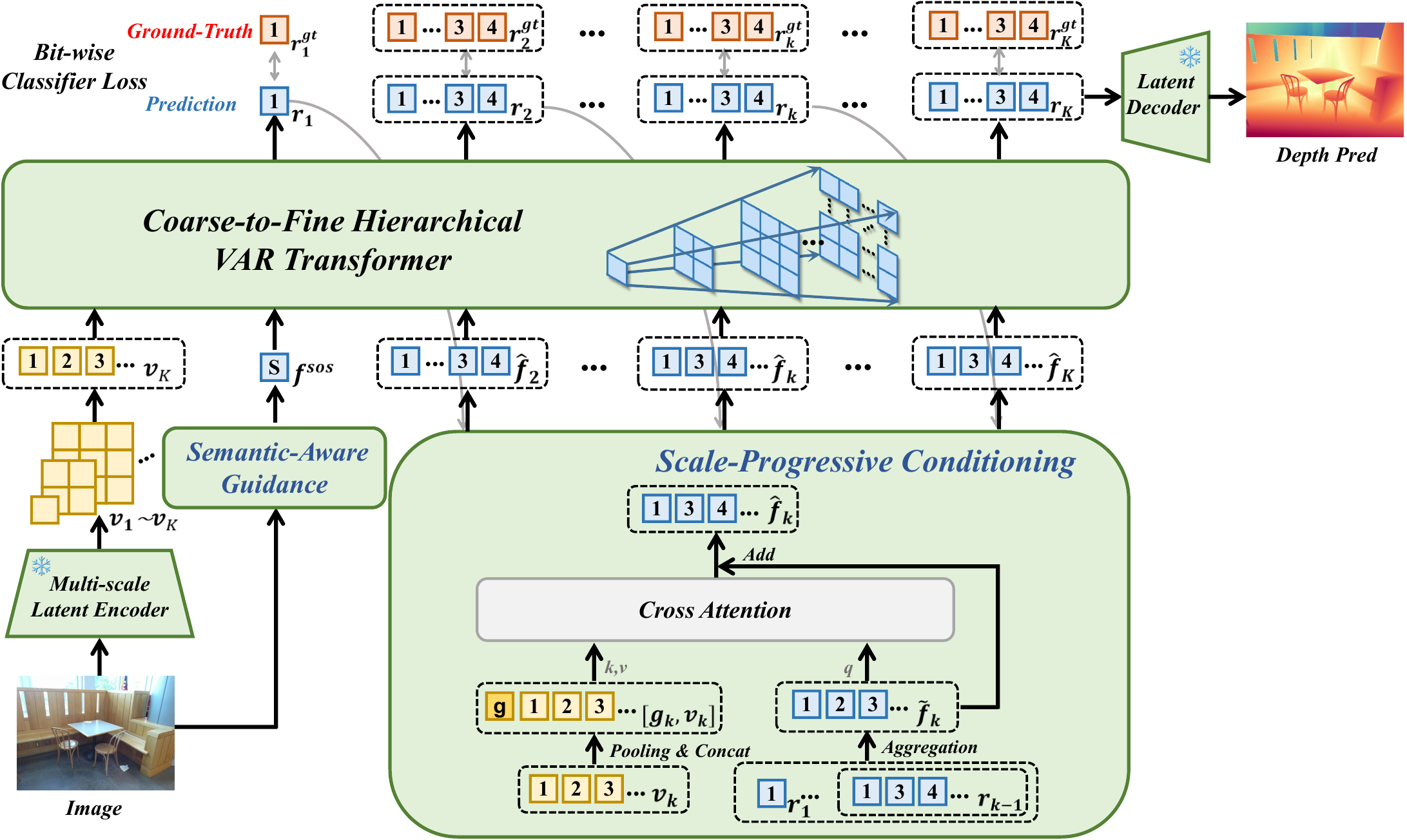}
    \caption{The overall framework of \methodName{}. \methodName{} formulates monocular depth estimation as a hierarchical auto-regressive generation process, where multi-scale residuals are predicted sequentially from coarse to fine. The visual tokenizer extracts multi-level image features, which are fused by the Scale-Progressive Conditioning (SPC) module to provide dense and spatially aligned conditioning for each generation stage. In parallel, the Semantic-Aware Guidance (SAG) module encodes high-level semantic cues from a VLM into an informative start-of-sequence token. The coarse-to-fine hierarchical VAR transformer integrates both sources of guidance to produce structured, globally consistent, and locally precise depth predictions, which are finally decoded into the full-resolution depth map.}
    \label{fig:model}
    \vspace{-0.6cm}
\end{figure*}


\section{Methodology}
In this section, we present \textbf{\methodName{}}, a hierarchical auto-regressive framework for monocular depth estimation. We begin by formalizing the task setting in Sec.~\ref{sec:method-pre}. Sec.~\ref{sec:method-arch} then describes the overall architecture of \methodName{}, including its training and inference pipelines, as well as the integration of conditioning signals. Sec.~\ref{sec:method-spc} details the hierarchical coarse-to-fine auto-regression process, with a focus on the \textbf{Scale-Progressive Conditioning (SPC)} module that injects structured multi-scale visual features at each generation stage to enhance local geometric fidelity. Finally, Sec.~\ref{sec:method-sag} introduces the \textbf{Semantic-Aware Guidance (SAG)} module, which provides high-level semantic priors that enforce global coherence and structural consistency across the predicted depth map.

\subsection{Task Formulation}
\label{sec:method-pre}
Given a single RGB image $\mathbf{I} \in \mathbb{R}^{H \times W \times 3}$, monocular depth estimation (MDE) aims to predict a dense depth map $\mathbf{D} \in \mathbb{R}^{H \times W}$, where each pixel value $\mathbf{D}_{i,j}$ represents the distance from the camera to the corresponding scene point. Formally, MDE seeks to learn a mapping function
\begin{equation}
f_{\boldsymbol{\theta}}: \mathbf{I} \rightarrow \mathbf{D},
\end{equation}
where $f_{\boldsymbol{\theta}}$ is parameterized by $\boldsymbol{\theta}$ and extracts geometric structure from monocular cues. Following prior works \cite{Marigold,depthfm,depthmaster}, we evaluate the \textit{zero-shot} MDE performance, where the model is expected to generalize its depth prediction capability to unseen domains or datasets without any task-specific fine-tuning.

\subsection{Overall Architecture}
\label{sec:method-arch}
We present \textbf{\methodName{}}, a hierarchical auto-regressive (AR) framework tailored for monocular depth estimation. The overall architecture is illustrated in Fig.~\ref{fig:model}. \methodName{} builds upon a pretrained VAR backbone, with a coarse-to-fine hierarchical VAR transformer \cite{tian2024visual} serving as its core component.

\paragraph{Training pipeline.}
Given an RGB image~$\mathbf{I}$, a descriptive text prompt~$\mathbf{T}$, and the ground-truth depth map~$\mathbf{D}$, the same tokenizer encodes both $\mathbf{I}$ and $\mathbf{D}$ into latent representations and decomposes them into multi-scale sequences of residual maps. These residuals serve as the visual conditioning inputs and auto-regressive supervision targets, respectively. During training, the transformer is trained under teacher forcing: the ground-truth residual at each scale is provided as context for predicting the next one. A block-wise causal attention mask enforces that the transformer only attends to residuals from preceding scales. Following Infinity~\cite{han2025infinity}, \methodName{} is optimized using a bit-wise classification loss with self-correction augmentation.

\paragraph{Inference pipeline.}
At inference time, depth supervision is not available and \methodName{} performs fully auto-regressive prediction. Residual maps are generated progressively from coarse to fine scales, with each predicted residual immediately incorporated into the context for subsequent stages. Once all residual levels are produced, they are fused by the tokenizer and decoded into the final full-resolution depth map. Unlike training, inference proceeds without any causal masking.

\methodName{} also strengthens conditioning signals for auto-regressive generation. In previous AR-based generative frameworks \cite{mao2025VAREdit,wang2025dar,gabdullin2024depthart}, conditioning is typically realized through global feature concatenation or shallow cross-attention between the visual/textual encoders and the transformer backbone. For the visual modality in particular, usually only the highest-resolution feature map from the visual encoder is flattened and used as the conditioning sequence, to limit the overall token length. Although computationally convenient, this design discards informative intermediate-scale features and weakens spatial coupling between conditioning signals and generated tokens, limiting local geometric accuracy. To overcome this limitation, \methodName{} introduces the \textbf{Scale-Progressive Conditioning (SPC)} module, which extracts multi-level visual features from the input image and injects them at the corresponding generation stage. This allows coarse scales to receive global structural cues and finer scales to access high-frequency visual evidence, improving spatial alignment and detail fidelity with minimal overhead. Details of SPC are provided in Sec.~\ref{sec:method-spc}. Complementary to SPC, \methodName{} integrates a \textbf{Semantic-Aware Guidance (SAG)} module that supplies scene-level semantic priors obtained from multi-modal language models. These global cues enhance the model’s understanding of layout and object relations, promoting coherent depth predictions across the entire scene. Further details of SAG are provided in Sec.~\ref{sec:method-sag}.

\subsection{Coarse-to-Fine Auto-regression}
\label{sec:method-spc}

At the core of \methodName{} is a hierarchical VAR transformer that predicts depth in a coarse-to-fine manner. 
During training, given an RGB image~$\mathbf{I}$, a textual description~$\mathbf{T}$, and a ground-truth depth map~$\mathbf{D}$, the multi-scale tokenizer encodes $\mathbf{D}$ into a latent tensor and decomposes it into a sequence of residual maps $\mathbf{R}=\{\mathbf{r}_1,\dots,\mathbf{r}_K\}$ with progressively increasing spatial resolution. 
This residual decomposition is important for dense geometric prediction. 
Coarse residuals encode global scene layout and dominant surface structures, while later residuals capture high-frequency corrections such as object boundaries, local discontinuities, and thin structures. 
Compared with predicting a full-resolution token sequence directly, this factorization reduces the modeling burden at each generation step and aligns the AR process with the scale-dependent complexity of depth maps. 
It also provides a natural interface for injecting visual conditions at matched spatial scales, which motivates the proposed Scale-Progressive Conditioning module.
The transformer models the joint distribution of these residual maps auto-regressively:
\begin{equation}
\begin{aligned}
    p(\mathbf{r}_1,\dots,\mathbf{r}_K)
    = \prod_{k=1}^{K}
    p(\mathbf{r}_k \mid \mathbf{r}_{<k},
    \psi_v(\mathbf{I}), \psi_t(\mathbf{T})),
\end{aligned}
\label{eq:var_likelihood}
\end{equation}
where $\psi_v$ and $\psi_t$ provide visual and semantic conditioning streams. At each step $k$, \methodName{} first constructs an intermediate representation $\mathbf{f}_k$ by aggregating all previously generated residuals:
\begin{equation}
    \mathbf{f}_k = \sum_{i=1}^{k-1} \mathrm{Up}(\mathrm{LookUp}(\mathbf{r}_i, \mathbf{C}), (h, w)),
\end{equation}
where $\mathrm{LookUp}(\cdot, \mathbf{C})$ retrieves vector embeddings from a learned codebook $\mathbf{C}$, and $\mathrm{Up}(\cdot, (h, w))$ denotes upsampling to a unified spatial size $h \times w$. The aggregated feature $\mathbf{f}_k$ is then downsampled to match the target resolution $(h_k, w_k)$:
\begin{equation}
\mathbf{\tilde{f}}_k = \mathrm{Down}(\mathbf{f}_k, (h_k, w_k)),
\end{equation}
where $\mathrm{Down}(\cdot, (h_k, w_k))$ represents bilinear downsampling that aligns the spatial size of the input $\mathbf{\tilde{f}}_k$ and the target $\mathbf{r}_k$ for the transformer at scale $k$. Following prior hierarchical AR paradigms \cite{han2025infinity, mao2025VAREdit}, the model operates under teacher forcing during training, while the generation process is initialized with $\mathbf{f}_1$, a start-of-sequence embedding derived from both visual and textual conditions. After obtaining all of the $K$ residual maps, they are aggregated to form the final feature map, which is passed to a latent decoder to synthesize the predicted depth map. Although the hierarchical auto-regressive process enables \methodName{} to refine depth from coarse structures to fine geometry, its effectiveness still hinges on strong conditioning signals at each scale. Standard AR models typically use only a single global visual feature as conditioning, which weakens the spatial alignment between conditioning cues and the predicted residuals. This limits the model's ability to recover fine structures and maintain geometric consistency. To address this issue, we introduce the \textbf{Scale-Progressive Conditioning (SPC)} module.

\paragraph{Scale-Progressive Conditioning.}
The Scale-Progressive Conditioning (SPC) module aims to provide dense, multi-scale visual conditioning for the transformer, improving local depth accuracy while maintaining global consistency. Formally, the SPC-guided feature at scale $k$ is defined as:
\begin{equation}
    \mathbf{\hat{f}}_k = \mathrm{SPC}(\mathbf{\tilde{f}}_k, \psi_v(\mathbf{I})),
\end{equation}
where $\mathbf{\hat{f}}_k$ is subsequently used as the transformer input for predicting $\mathbf{r}_k$. To be specific, let $\mathbf{v}_k \in \mathbb{R}^{L_k \times c}$ denote the local feature tokens at scale $k$ and $\mathbf{g}_k \in \mathbb{R}^{1 \times c}$ the pooled global context. $\mathbf{\tilde{f}}_k \in \mathbb{R}^{L_k \times c}$ is the intermediate sequence of predicted residuals at the same scale. SPC refines $\mathbf{\tilde{f}}_k$ by attending to both local and global features:
\begin{equation}
\mathbf{Q}_k = \mathbf{\tilde{f}}_k \mathbf{W}_Q^\top, \; 
\mathbf{K}_k = [\mathbf{g}_k; \mathbf{v}_k] \mathbf{W}_K^\top, \; 
\mathbf{V}_k = [\mathbf{g}_k; \mathbf{v}_k] \mathbf{W}_V^\top,
\end{equation}
where $\mathbf{W}_Q, \mathbf{W}_K, \mathbf{W}_V \in \mathbb{R}^{d \times c}$ project input features of dimension $c$ to attention dimension $d$, and $[\cdot;\cdot]$ denotes concatenation along the sequence dimension. Here, $L_k$ is the sequence length of $\mathbf{\tilde{f}}_k$ and $\mathbf{v}_k$. The SPC-enhanced feature at scale $k$ is then computed as:
\begin{equation}
\mathbf{\hat{f}}_k = \mathbf{\tilde{f}}_k + 
\mathrm{softmax}\Big(\frac{\mathbf{Q}_k \mathbf{K}_k^\top}{\sqrt{d}}\Big) \mathbf{V}_k,
\end{equation}
resulting in $\mathbf{\hat{f}}_k \in \mathbb{R}^{L_k \times c}$. This formulation allows each intermediate token to attend adaptively to both fine-grained local cues and the global scene feature, preserving local details while maintaining global geometric consistency. This design enables adaptive conditioning at each scale while preserving computational efficiency. We provide detailed ablation studies and analysis of the SPC design in Sec.~\ref{sec:exp-ablation} to validate its effectiveness and each design choice.

\begin{figure*}[!ht]
    \centering
    \includegraphics[width=\linewidth]{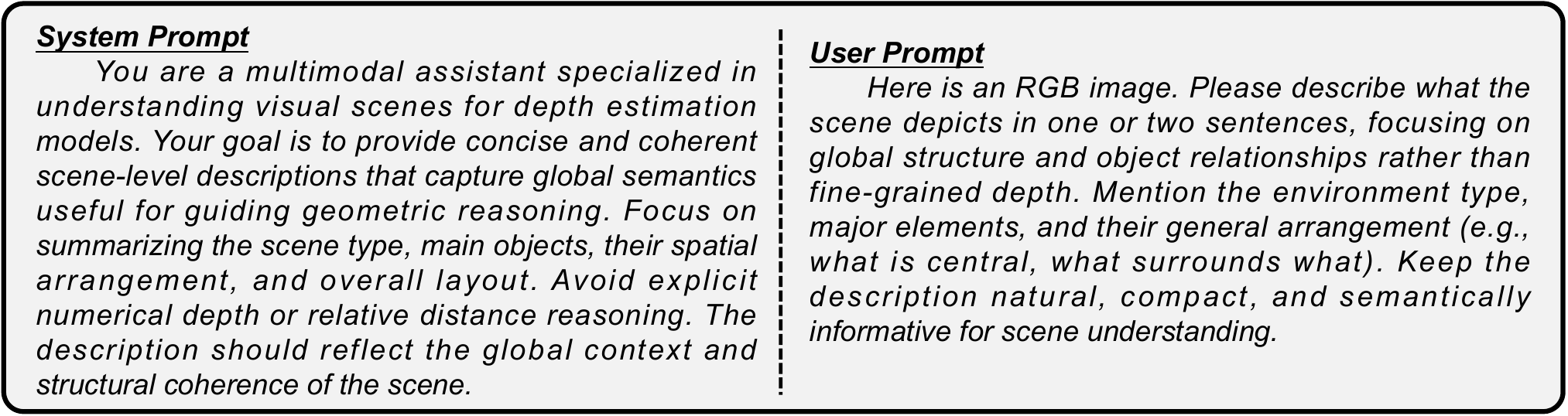}
    \caption{System and user prompts provided to the multi-modal language model (VLM) in the SAG module. Instead of guiding the VLM to describe explicit depth details, these prompts encourage the model to convey high-level semantic and structural understanding of the scene, providing global contextual cues that enhance consistency in the predicted depth map of \methodName{}.}
    \label{fig:prompt}
\end{figure*}

\subsection{Semantic-Aware Guidance}
\label{sec:method-sag}
Furthermore, to incorporate scene-level semantics, we introduce the Semantic-Aware Guidance (SAG) module, which provides high-level textual priors as additional conditioning for the auto-regressive depth generation process. Specifically, given an input RGB image $\mathbf{I}$ and a textual prompt $\mathbf{P}$ (illustrated in Fig.~\ref{fig:prompt}), a multi-modal language model (VLM) generates a descriptive text $\mathbf{T}'$ that summarizes the scene and captures object relations. To explicitly guide depth estimation, we append a fixed instruction, e.g., ``Generate the depth map of this RGB image.", forming the final textual input $\mathbf{T}$:
\begin{equation}
\mathbf{T} = \text{Concat}(\mathbf{T}', \text{instruction}).
\end{equation}
Following prior AR-based frameworks \cite{mao2025VAREdit, han2025infinity}, $\mathbf{T}$ is encoded by a text encoder (e.g., Flan-T5 \cite{chung2024flant5}) into embeddings:
\begin{equation}
\mathbf{t} = \phi_t(\mathbf{T}), \quad \mathbf{t} \in \mathbb{R}^{L \times d_t},
\end{equation}
where $L$ is the text length and $d_t$ denotes the embedding dimension. These embeddings are then projected and aggregated into a learnable start-of-sequence (sos) token:
\begin{equation}
\mathbf{f}^{\mathrm{sos}} = \mathbf{W}_T \mathbf{t},
\end{equation}
where $\mathbf{f}^{\mathrm{sos}} \in \mathbb{R}^{1 \times 1 \times d}$ and serves as the initial input for the first prediction scale of the transformer. This design allows the model to leverage both scene-specific textual cues generated from the image and the instructional prompt, providing strong semantic guidance from the very first generation step. By combining SAG with SPC, \methodName{} unifies dense visual conditioning and global semantic guidance, enabling fine-grained local depth predictions that are consistent with the overall scene structure. Ablation studies in Sec.~\ref{sec:exp-ablation} explore the contribution of SAG and different textual integration strategies.

\section{Experiments}
\subsection{Experimental Setup}

\paragraph{Training datasets}
To train the proposed \methodName{}, we utilize two synthetic datasets, Hypersim \cite{roberts2021hypersim} and Virtual KITTI~\cite{vkitti}, covering both indoor and outdoor scenes. Following previous works~\cite{Marigold,depthfm,depthmaster}, we sample 54K images from Hypersim and 20K from Virtual KITTI for training. To align with training protocols adopted by some of the prior works \cite{he2024lotus,gabdullin2024depthart}, we also train a model variant using only the 54K Hypersim samples. In addition, to further assess how \methodName{} performs under larger training regimes, we construct an expanded dataset by sampling an additional 100K images from several publicly available sources, including UrbanSyn~\cite{gomez2025Urbansyn}, UnrealStereo4K~\cite{UnrealStereo4K}, TartanAir~\cite{wang2020tartanair}, DDAD~\cite{DDAD} and SA-1B~\cite{SegmentAnything}. For samples with reliable depth annotations, we directly use the provided ground truth. For samples without depth labels or with lower-quality annotations, we follow the paradigm established by the Depth Anything family~\cite{DA,DA2} and generate pseudo ground truth. This enlarged training corpus enables us to examine the scalability of our hierarchical auto-regressive framework and to verify whether \methodName{} can benefit simultaneously from pretrained priors and data scaling. The impact of this expanded training corpus is analyzed in Sec.~\ref{sec:exp-sota}.

\begin{table*}[!ht]
\centering
\caption{Zero-shot monocular depth estimation across multiple benchmarks. Better: AbsRel$\downarrow$, $\delta_1\uparrow$. \textbf{Bold} numbers indicate the best result within each paradigm. \methodName{} rows are highlighted with a background color.}
\scriptsize
\resizebox{\textwidth}{!}{
\begin{tabular}{c|c|cc|cc|cc|cc|cc}
\toprule
\multirow{2}{*}{Method} & Training & \multicolumn{2}{c|}{NYUv2} & \multicolumn{2}{c|}{KITTI} & \multicolumn{2}{c|}{ETH3D} & \multicolumn{2}{c|}{ScanNet} & \multicolumn{2}{c}{DIODE} \\ \cline{3-12}
~ & Data & AbsRel$\downarrow$ & $\delta_1\uparrow$ & AbsRel$\downarrow$ & $\delta_1\uparrow$ & AbsRel$\downarrow$ & $\delta_1\uparrow$ & AbsRel$\downarrow$ & $\delta_1\uparrow$ & AbsRel$\downarrow$ & $\delta_1\uparrow$ \\
\midrule
\multicolumn{12}{c}{\textit{\textbf{Discriminative / Traditional Data-driven Model}}} \\
\midrule
DiverseDepth~\cite{yin2020diversedepth} & 320K & 11.7 & 87.5 & 19.0 & 70.4 & 22.8 & 69.4 & 10.9 & 88.2 & - & - \\
MiDaS~\cite{MiDas} & 2M & 11.1 & 88.5 & 23.6 & 63.0 & 18.4 & 75.2 & 12.1 & 84.6 & - & - \\
LeReS~\cite{yin2021learning} & 354K & 9.0 & 91.6 & 14.9 & 78.4 & 17.1 & 77.7 & 9.1 & 91.7 & - & - \\
Omnidata~\cite{eftekhar2021omnidata} & 12M & 7.4 & 94.5 & 14.9 & 83.5 & 16.6 & 77.8 & 7.5 & 93.6 & - & - \\
HDN~\cite{zhang2022hierarchical} & 300K & \textbf{6.9} & \textbf{94.8} & 11.5 & 86.7 & 12.1 & 83.3 & \textbf{8.0} & \textbf{93.9} & - & - \\
DPT~\cite{DPT} & 1.2M & 9.8 & 90.3 & \textbf{10.0} & \textbf{90.1} & \textbf{7.8} & \textbf{94.6} & 8.2 & 93.4 & - & - \\
\midrule
\multicolumn{12}{c}{\textit{\textbf{Depth Anything (DA) Family}}} \\
\midrule
DA~\cite{DA} & 62M & \textbf{4.3} & \textbf{98.1} & 7.6 & \textbf{94.7} & \textbf{12.7} & \textbf{88.2} & - & - & \textbf{6.6} & \textbf{95.2} \\
DA2~\cite{DA2} & 62M & 4.5 & 97.9 & \textbf{7.4} & 94.6 & 13.1 & 86.5 & \textbf{6.5} & \textbf{97.2} & \textbf{6.6} & \textbf{95.2} \\
\midrule
\multicolumn{12}{c}{\textit{\textbf{Diffusion Prior Based Model}}} \\
\midrule
PriorDiffusion~\cite{zeng2024priordiffusion} & 74K & 5.9 & 96.0 & 10.4 & 90.3 & 6.5 & 95.8 & 6.7 & 94.9 & - & - \\
Marigold~\cite{Marigold} & 74K & 5.5 & 96.4 & 9.9 & 91.6 & 6.5 & 96.0 & 6.4 & 95.1 & 10.0 & 90.7 \\
GeoWizard~\cite{fu2024geowizard} & 28K & 5.2 & 96.6 & 9.7 & 92.1 & 6.4 & 96.1 & 6.1 & 95.3 & 12.0 & 89.8 \\
DepthFM~\cite{depthfm} & 74K & 5.5 & 96.3 & 8.9 & 91.3 & 5.8 & 96.2 & 6.3 & 95.4 & - & - \\
GenPercept~\cite{GenPercept} & 90K & 5.2 & 96.6 & 9.4 & 92.3 & 6.6 & 95.7 & 5.6 & 96.5 & - & - \\
Lotus~\cite{he2024lotus} & 54K & 5.4 & 96.8 & 8.5 & 92.2 & 5.9 & 97.0 & 5.9 & 95.7 & \textbf{9.8} & \textbf{92.4} \\
DepthMaster~\cite{depthmaster} & 74K & \textbf{5.0} & \textbf{97.2} & \textbf{8.2} & \textbf{93.7} & \textbf{5.3} & \textbf{97.4} & \textbf{5.5} & \textbf{96.7} & 21.5 & 77.6\\
\midrule
\multicolumn{12}{c}{\textit{\textbf{Auto-Regressive Paradigms}}} \\
\midrule
VAR~\cite{tian2024visual} & 54K & 14.1 & 81.5 & - & - & 28.5 & 75.5 & - & - & - & - \\
DepthART~\cite{gabdullin2024depthart} & 54K & 11.5 & 85.9 & 18.1 & 70.3 & 17.7 & 80.4 & 16.0 & 78.0 & 29.8 & 69.0 \\
\rowcolor{gray!15} \methodName{} (Ours) & 54K & 5.0 & 97.0 & 8.9 & 92.7 & 5.2 & 97.2 & 5.3 & 96.5 & 7.8 & 94.0 \\
\rowcolor{gray!15} \methodName{} (Ours) & 74K & 4.8 & 97.5 & 8.4 & 93.5 & 4.9 & 97.6 & 5.0 & 97.0 & 6.9 & 95.4 \\
\rowcolor{gray!15} \methodName{} (Ours) & 174K & \textbf{4.5} & \textbf{97.7} & \textbf{7.7} & \textbf{94.0} & \textbf{4.7} & \textbf{97.7} & \textbf{4.8} & \textbf{97.2} & \textbf{6.5} & \textbf{95.7} \\
\bottomrule
\end{tabular}
}
\label{tab:sota}
\vspace{-0.4cm}
\end{table*}

\paragraph{Evaluation datasets and metrics}
Following prior works~\cite{Marigold,depthfm,depthmaster}, zero-shot performance is evaluated on five real-world benchmarks, including NYUv2~\cite{nyuv2}, KITTI~\cite{kitti}, ETH3D~\cite{eth3d}, ScanNet~\cite{dai2017scannet}, and DIODE~\cite{vasiljevic2019diode}. These datasets cover diverse indoor and outdoor scenarios. To be specific, NYUv2 and ScanNet contain indoor RGB-D scenes captured by Kinect-style sensors, KITTI provides outdoor street scenes with sparse LiDAR depth, while ETH3D and DIODE provide high-resolution images with LiDAR-derived depth maps.
Depth prediction quality is measured by Absolute Relative Error (AbsRel) and $\delta_1$ accuracy. AbsRel is defined as
\begin{equation}
\mathrm{AbsRel} =
\frac{1}{HW}
\sum_{i=1}^{H}
\sum_{j=1}^{W}
\frac{
\left|D_{\mathrm{gt}}^{(i,j)} - D_{\mathrm{pred}}^{(i,j)}\right|
}{
D_{\mathrm{gt}}^{(i,j)}
},
\end{equation}
where $D_{\mathrm{gt}}$ and $D_{\mathrm{pred}}$ denote the ground-truth and predicted depth maps. The $\delta_1$ accuracy measures the fraction of pixels where the prediction is within a threshold of the ground truth:
\begin{equation}
\delta_1 = \frac{1}{HW} \sum_{i=1}^{H}\sum_{j=1}^{W} \big[\max\big(\frac{D_{pred}^{(i,j)}}{D_{gt}^{(i,j)}}, \frac{D_{gt}^{(i,j)}}{D_{pred}^{(i,j)}}\big) < 1.25\big],
\end{equation}
where $[\cdot]$ denotes the Iverson bracket. By definition, lower AbsRel values indicate better accuracy, whereas higher $\delta_1$ values correspond to more precise depth predictions.
In addition, we evaluate relative depth estimation on DA-2K~\cite{DA2}, a benchmark introduced by Depth Anything V2. DA-2K contains 1K images with 2K pairwise relative depth annotations across eight diverse scenarios. Following the official protocol, we report Accuracy (Acc.), which measures the percentage of annotated pixel pairs whose relative depth ordering is correctly predicted.

\paragraph{Depth normalization}
Following common practice in monocular depth estimation~\cite{MiDas,Marigold}, raw depth maps are linearly normalized into the range $[-1,1]$ before training. For each depth map $D$, the normalized ground-truth depth $D_{\mathrm{gt}}$ is computed as
\begin{equation}
    D_{\mathrm{gt}} =
    \left(
    \frac{D - D_{2}}{D_{98} - D_{2}} - \frac{1}{2}
    \right) \times 2,
\end{equation}
where $D_{2}$ and $D_{98}$ denote the $2\%$ and $98\%$ percentiles of valid depth values, respectively. This percentile-based normalization suppresses the influence of extreme outliers and encourages the model to learn affine-invariant depth structure rather than dataset-specific metric scale.

\paragraph{Implementation details}
\methodName{} is initialized from the pretrained 8B weights of VAREdit~\cite{mao2025VAREdit}. A positional offset of $\Delta=(64,64)$ is applied to the 2D Rotary Position Embeddings (2D-RoPE) of all conditioning tokens to distinguish source and target image tokens. Qwen2.5-VL-7B-Instruct~\cite{qwen2.5-VL} is used as the pretrained VLM in the SAG module.
Training is conducted with PyTorch on eight NVIDIA A800 GPUs for $\sim$60K iterations, using a batch size of 4 and an initial learning rate of $5 \times 10^{-5}$. During inference, classifier-free guidance is applied with a guidance scale of $\eta=3$, and the logits temperature is set to $\tau=0.5$. KV caching is enabled to accelerate auto-regressive decoding.
To maintain a consistent token length while supporting variable input resolutions, each image is first resized so that its longer side is 1024 pixels while preserving the aspect ratio. The shorter side is then padded to 1024 pixels. The predicted depth map is finally resized back to the original image resolution for the final output.

\begin{table}[!ht]
    \centering
    \caption{Comparison with previous methods on DA-2K. \methodName{} rows are highlighted with a background color.}
    \resizebox{\columnwidth}{!}{
    \begin{tabular}{c|c|c}
    \toprule
    Method & Training Data & Accuracy ($\%$) $\uparrow$ \\
    \midrule
    DepthART~\cite{gabdullin2024depthart} & 54K & 72.9 \\
    DepthFM~\cite{depthfm} & 74K & 85.8 \\
    Marigold~\cite{Marigold} & 74K & 86.8 \\
    GeoWizard~\cite{fu2024geowizard} & 28K & 88.1 \\
    DA~\cite{DA} & 62M & 88.5 \\
    \rowcolor{gray!15} \methodName{} (Ours) & 54K & \textbf{95.1} \\
    DA2-ViT-S~\cite{DA2} & 62M & 95.3 \\
    \rowcolor{gray!15} \methodName{} (Ours) & 74K & \textbf{95.8} \\
    \rowcolor{gray!15} \methodName{} (Ours) & 174K & \textbf{96.7} \\
    DA2-ViT-M~\cite{DA2} & 62M & 97.0 \\
    DA2-ViT-L~\cite{DA2} & 62M & 97.1 \\
    DA2-ViT-G~\cite{DA2} & 62M & 97.4 \\ 
    \bottomrule
    \end{tabular}
    }
    \label{tab:da2k}
\end{table}

\subsection{Quantitative Comparison with State-of-the-Arts}
\label{sec:exp-sota}

\paragraph{Zero-shot monocular depth estimation}
Table~\ref{tab:sota} reports zero-shot monocular depth estimation results on NYUv2, KITTI, ETH3D, ScanNet, and DIODE.
The compared methods cover three representative paradigms: discriminative data-driven models, diffusion-prior-based models, and auto-regressive models.
This comparison allows us to evaluate \methodName{} not only against individual methods, but also across the main modeling paradigms used in recent monocular depth estimation.

Under the 74K training setting, \methodName{} achieves strong performance across all five benchmarks.
Compared with diffusion-prior-based methods trained under similar data scales, \methodName{} obtains consistently competitive or superior results.
For example, compared with Marigold~\cite{Marigold}, \methodName{} improves AbsRel from 5.5 to 4.8 on NYUv2, from 9.9 to 8.4 on KITTI, from 6.5 to 4.9 on ETH3D, from 6.4 to 5.0 on ScanNet, and from 10.0 to 6.9 on DIODE.
These results suggest that hierarchical auto-regressive generation can provide an explicit coarse-to-fine alternative to diffusion-based refinement for structured geometric prediction.
Rather than recovering depth through iterative global denoising, \methodName{} constructs depth from coarse layout to fine structures, which better matches the multi-scale organization of scene geometry.

\paragraph{Comparison with auto-regressive baselines}
The advantage of \methodName{} is particularly clear when compared with existing auto-regressive baselines.
Directly applying generic visual AR models to depth prediction, as in VAR~\cite{tian2024visual} and DepthART~\cite{gabdullin2024depthart}, leads to substantially weaker performance.
This gap indicates that auto-regression alone is insufficient for dense geometric estimation.
Depth prediction requires not only sequential generation, but also a task-aware factorization that preserves spatial alignment, local detail, and global geometric consistency.
The results in Table~\ref{tab:sota} support this conclusion: \methodName{} substantially narrows the gap between auto-regressive modeling and state-of-the-art diffusion-based depth estimation, while further surpassing many diffusion-prior-based methods under comparable data scales.

\begin{table}[!ht]
    \centering
    \caption{Additional in-domain comparison with DAR on NYUv2 under the training protocol reported by DAR~\cite{wang2025dar}.}
    \label{tab:indomain}
    \begin{tabular}{l|c}
    \toprule
    Method & AbsRel$\downarrow$ \\
    \midrule
    DAR~\cite{wang2025dar} & 4.4 \\
    ARDepth-2B & 3.9 \\
    ARDepth-8B & \textbf{3.5} \\
    \bottomrule
    \end{tabular}
\end{table}

\paragraph{Additional in-domain comparison with DAR}
DAR~\cite{wang2025dar} is a closely related auto-regressive depth estimation method, but its reported results mainly follow an in-domain NYUv2 protocol rather than the zero-shot evaluation setting used in Table~\ref{tab:sota}. 
To provide a more direct comparison with this AR-depth baseline, we additionally evaluate \methodName{} under the same in-domain protocol reported by DAR. 
Since DAR mainly reports AbsRel under this setting, we use AbsRel as the primary metric for this additional comparison.
As shown in Table~\ref{tab:indomain}, \methodName{} achieves an AbsRel of 3.5 with the 8B model and 3.9 with the parameter-aligned 2B variant, both outperforming DAR's reported AbsRel of 4.4.
This result complements the zero-shot comparisons above. 
It suggests that the advantage of \methodName{} does not only come from cross-domain generalization, but also holds under an in-domain setting against a directly related AR-based depth model. 
The improvement further supports the importance of scale-progressive generation and stage-aligned conditioning, beyond the general adoption of auto-regressive modeling.

\paragraph{Data scaling and relative depth estimation}
We further examine whether \methodName{} can benefit from additional training data.
In addition to the default 74K training set, we train a 54K variant using only Hypersim and a 174K variant using the expanded training corpus.
The 54K variant already achieves strong zero-shot generalization, outperforming prior AR methods and remaining competitive with diffusion-based methods.
When the training set is increased from 74K to 174K samples, \methodName{} obtains consistent improvements across all evaluation datasets.
Specifically, AbsRel improves from 4.8 to 4.5 on NYUv2, from 8.4 to 7.7 on KITTI, from 4.9 to 4.7 on ETH3D, from 5.0 to 4.8 on ScanNet, and from 6.9 to 6.5 on DIODE.
These improvements show that the proposed framework is not only data-efficient at moderate training scales, but also scalable when additional supervision is available.

Table~\ref{tab:da2k} further evaluates relative depth estimation on DA-2K.
This benchmark focuses on pairwise depth ordering and therefore provides a complementary view beyond metric depth errors.
The 74K version of \methodName{} achieves 95.8\% accuracy, surpassing diffusion-based counterparts such as Marigold~\cite{Marigold}, DepthFM~\cite{depthfm}, and GeoWizard~\cite{fu2024geowizard}.
It also surpasses DA2-ViT-S, despite DA2 being trained on a substantially larger corpus.
With 174K training samples, \methodName{} further improves to 96.7\%, approaching the performance of larger DA2 variants.
This result suggests that \methodName{} does not merely improve local boundary sharpness, but also preserves coherent depth ordering across objects and regions.
Overall, the quantitative results show that hierarchical auto-regressive generation is competitive, data-efficient, and scalable for dense geometric prediction.




\subsection{Qualitative Comparison}
\label{sec:exp-qualitative}

\begin{figure*}[!t]
    \centering
    \includegraphics[width=\linewidth]{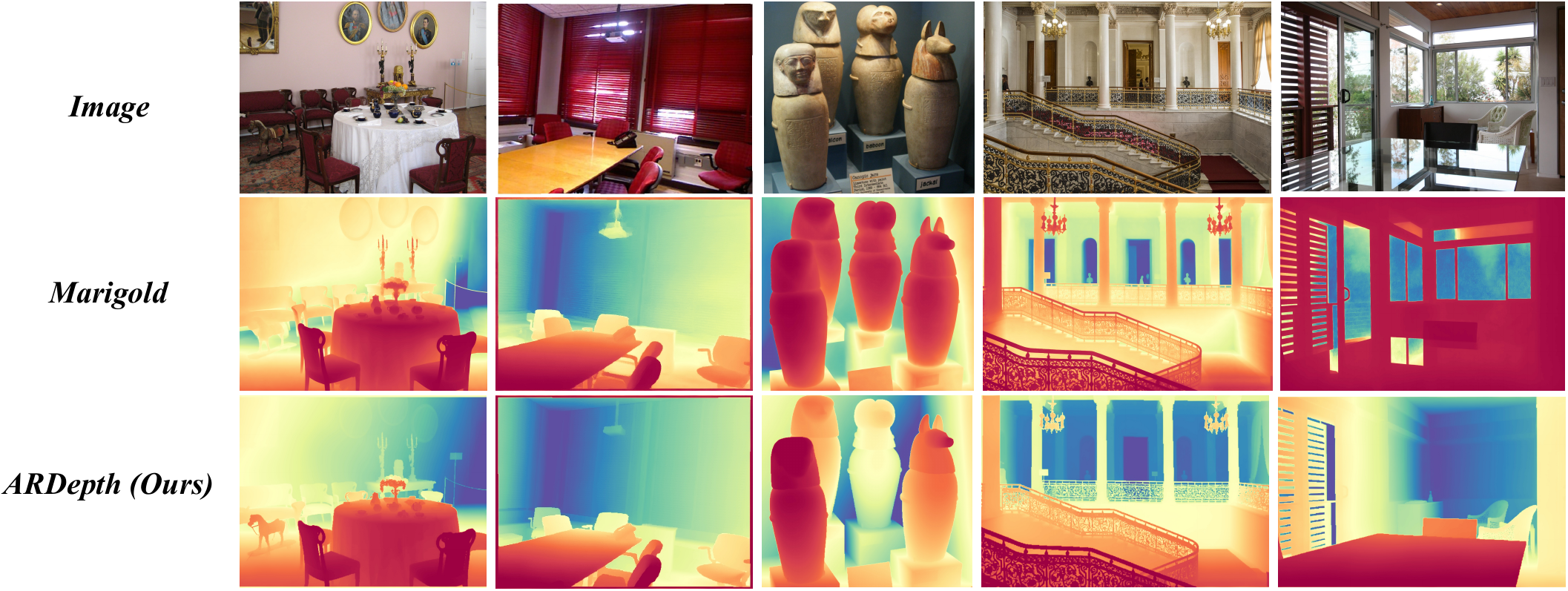}
    \caption{Qualitative comparison with Marigold~\cite{Marigold}. \methodName{} better preserves object boundaries, thin structures, and local depth discontinuities while maintaining coherent global layout, illustrating the benefit of coarse-to-fine auto-regressive depth generation.}
    \label{fig:vis}
\end{figure*}

Beyond numerical performance, Fig.~\ref{fig:vis} provides qualitative comparisons with Marigold~\cite{Marigold}.
In challenging indoor scenes with furniture, windows, thin objects, and layered surfaces, \methodName{} produces depth maps with clearer object boundaries and more faithful local discontinuities.
For example, foreground objects are better separated from background regions, and depth transitions around structural edges are sharper.
At the same time, the predictions maintain coherent global layout instead of fragmenting local structures.
These visual results are consistent with the quantitative improvements.
The coarse-to-fine AR formulation first establishes global scene layout and then progressively refines local details, while SPC injects scale-aligned visual cues during generation.
As a result, \methodName{} can better preserve fine structures without sacrificing global geometric consistency.

\subsection{Ablation Analysis}
\label{sec:exp-ablation}

We conduct ablation studies to examine whether the gains of \methodName{} come from the proposed task-aware design rather than from simply adopting a pretrained AR backbone.
The analysis covers four aspects: scale-progressive visual conditioning, semantic-aware textual guidance, hierarchical auto-regression, and AR resolution.
Together, these experiments evaluate the two central claims of \methodName{}: dense depth generation benefits from a coarse-to-fine AR structure, and each generation stage should receive visual or semantic conditions matched to its role in the hierarchy.

\begin{table}[!ht]
    \centering
    \caption{Ablation analysis of the SPC module. ``No Global Context'' removes the pooled global feature $\mathbf{g}_k$, while ``Direct Add (x)'' fuses multi-scale features directly with a scalar weight $x$. ``Full SPC'' employs attention-based fusion of local and global cues, achieving the best balance of local detail and global consistency.}
    \label{tab:spc}
    \resizebox{\columnwidth}{!}{
    \begin{tabular}{c|cc|cc}
    \toprule
    \multirow{2}{*}{Method} 
    & \multicolumn{2}{c|}{NYUv2} 
    & \multicolumn{2}{c}{KITTI} \\ 
    \cline{2-5}
    & AbsRel$\downarrow$ & $\delta_1 \uparrow$ 
    & AbsRel$\downarrow$ & $\delta_1 \uparrow$ \\
    \midrule
    No SPC & 6.0 & 95.1 & 9.5 & 92.1 \\
    Global Context Only & 5.6 & 95.9 & 9.1 & 92.4 \\
    No Global Context & 5.1 & 97.1 & 8.7 & 93.0 \\
    Direct Add (0.1) & 5.2 & 97.0 & 8.7 & 93.0 \\
    Direct Add (0.3) & 5.1 & 97.1 & 8.6 & 93.2 \\
    Direct Add (0.5) & 5.3 & 96.9 & 8.8 & 92.7 \\
    Full SPC & \textbf{4.8} & \textbf{97.5} & \textbf{8.4} & \textbf{93.5} \\
    \bottomrule
    \end{tabular}
    }
\end{table}

\paragraph{Ablation on SPC}
\label{sec:exp-spc}
We evaluate several variants to assess the contributions of key design choices in the SPC module:
(1) \textbf{No SPC}, where the module is entirely removed;
(2) \textbf{Global Context Only}, where only the pooled global context feature $\mathbf{g}_k$ is used;
(3) \textbf{No Global Context}, which uses only multi-scale local features $\mathbf{v}_k$ and discards $\mathbf{g}_k$;
(4) \textbf{Direct Add}, where multi-scale local features are directly added to the corresponding intermediate representations $\mathbf{\tilde{f}}_k$ with fixed scalar weights of 0.1, 0.3, and 0.5;
and (5) \textbf{Full SPC}, the default configuration, which adaptively fuses local and global cues through attention.
As shown in Tab.~\ref{tab:spc}, removing SPC leads to clear degradation on both NYUv2 and KITTI, showing that the original AR conditioning is insufficient for dense geometric prediction.
Using only global context improves over \textbf{No SPC}, but still lacks spatially aligned evidence for recovering boundaries and local structures.
In contrast, \textbf{No Global Context} performs much better, indicating that multi-scale local features play a more direct role in dense depth estimation.
This is consistent with the nature of MDE, where fine structures, object contours, and local discontinuities must be grounded in spatially aligned image evidence.
The \textbf{Direct Add} variants further show that simply injecting multi-scale features is not enough.
A moderate weight works reasonably well, while a large weight degrades performance, suggesting that uncontrolled feature injection may disturb the intermediate AR representation.
By contrast, \textbf{Full SPC} achieves the best performance by adaptively balancing local and global cues.
These results indicate that the gain of SPC does not merely come from using more visual features, but from stage-aligned and adaptive conditioning that allows each AR stage to access visual evidence at a suitable spatial scale.


\begin{table}[!ht]
    \centering
    \caption{Ablation study of textual context in SAG. \textbf{SAG Text} yields the best performance, highlighting the importance of compact scene-level semantic guidance for monocular depth estimation.}
    \label{tab:sag}
    \resizebox{0.92\columnwidth}{!}{
    \begin{tabular}{c|cc|cc}
    \toprule
    \multirow{2}{*}{Method} 
    & \multicolumn{2}{c|}{NYUv2} 
    & \multicolumn{2}{c}{KITTI} \\ 
    \cline{2-5}
    & AbsRel$\downarrow$ & $\delta_1 \uparrow$ 
    & AbsRel$\downarrow$ & $\delta_1 \uparrow$ \\
    \midrule
    Blank Text & 5.3 & 96.7 & 8.9 & 92.5 \\
    Unified Text & 5.1 & 97.0 & 8.6 & 92.8 \\
    SAG Text & \textbf{4.8} & \textbf{97.5} & \textbf{8.4} & \textbf{93.5} \\
    SAG Text Spatial & 4.9 & 97.3 & 8.6 & 93.2 \\
    \bottomrule
    \end{tabular}
    }
\end{table}

\paragraph{Ablation on textual guidance}
\label{sec:exp-sag}
We evaluate four textual-conditioning variants to analyze the role of SAG:
(1) \textbf{Blank Text}, where no textual input is provided;
(2) \textbf{Unified Text}, where all samples use the same task prompt ``Generate the depth map of this RGB image.'';
(3) \textbf{SAG Text}, where the VLM generates an image-specific scene description that is concatenated with the task prompt;
and (4) \textbf{SAG Spatial Text}, where the VLM is further prompted to describe object arrangements and relative depth relationships.
As shown in Tab.~\ref{tab:sag}, \textbf{Unified Text} improves over \textbf{Blank Text}, indicating that even a generic task instruction can help stabilize conditional generation.
Replacing the unified prompt with image-specific \textbf{SAG Text} further improves the results, showing that compact scene-level semantics provide useful global context for depth prediction.
This supports the motivation of SAG: textual guidance should provide semantic priors that regularize the overall scene layout, rather than directly predict pixel-level geometry.
Interestingly, \textbf{SAG Spatial Text} gives comparable but slightly worse performance than \textbf{SAG Text}.
A likely reason is that detailed spatial descriptions from off-the-shelf VLMs may contain inaccurate object relations or overly specific depth claims, which introduce noise into the generation process.
This result suggests that VLM-generated text is most useful as compact semantic context, rather than as explicit fine-grained geometric instruction.
Therefore, \methodName{} uses text to regularize global scene structure, while local geometric reconstruction remains grounded in visual conditioning.

\begin{figure}[!ht]
    \centering
    \includegraphics[width=\linewidth]{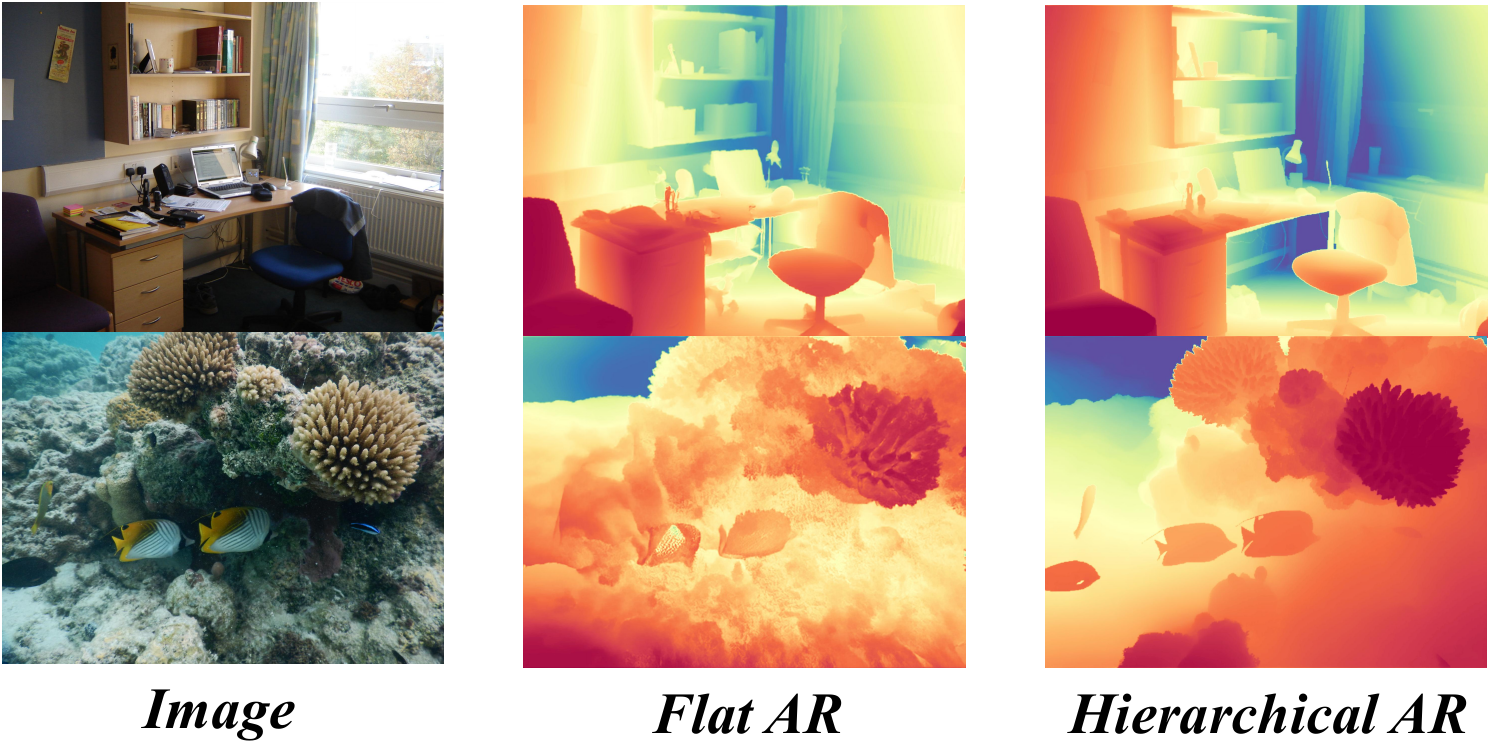}
    \caption{Qualitative ablation on auto-regression type. The flat AR baseline captures coarse layout but loses fine-grained depth variations, while the hierarchical AR formulation better preserves local structures and depth discontinuities.}
    \label{fig:vis2}
\end{figure}

\begin{table}[!ht]
    \centering
    \caption{Ablation study of the auto-regression type. The hierarchical AR formulation substantially outperforms the flat AR baseline, showing the importance of coarse-to-fine depth generation.}
    \label{tab:ar-type}
    \resizebox{0.82\columnwidth}{!}{
    \begin{tabular}{c|cc|cc}
    \toprule
    \multirow{2}{*}{Method} 
    & \multicolumn{2}{c|}{NYUv2} 
    & \multicolumn{2}{c}{KITTI} \\ 
    \cline{2-5}
    & AbsRel$\downarrow$ & $\delta_1 \uparrow$ 
    & AbsRel$\downarrow$ & $\delta_1 \uparrow$ \\
    \midrule
    Flat AR & 23.9 & 85.3 & 25.6 & 81.8 \\
    Hierarchical AR & \textbf{4.8} & \textbf{97.5} & \textbf{8.4} & \textbf{93.5} \\
    \bottomrule
    \end{tabular}
    }
\end{table}

\paragraph{Ablation on auto-regression type}
We evaluate the impact of hierarchical auto-regression by comparing our coarse-to-fine design with a single-stage flat AR baseline following the raster-scan design of Editar~\cite{mu2025editar}.
For a controlled comparison, the flat AR baseline operates at $512 \times 512$ resolution and predicts 1024 tokens in a single auto-regressive sequence.
Its parameter scale is kept comparable to \methodName{}, and it is trained under the same optimization protocol.
As illustrated in Fig.~\ref{fig:vis2}, the flat AR variant captures the general spatial layout and object placement, but struggles to recover fine-grained depth variations and sharp structures.
This limitation is quantitatively confirmed in Tab.~\ref{tab:ar-type}, where flat AR obtains substantially higher AbsRel and lower $\delta_1$ than the hierarchical formulation.
These results indicate that auto-regression alone is insufficient for dense geometric prediction.
In a flat raster-scan formulation, global layout, local discontinuities, and high-resolution details must be modeled within one long sequential process, which places a heavy burden on the AR model.
In contrast, \methodName{} factorizes depth generation into multiple coarse-to-fine stages.
Coarse stages establish the global layout and dominant surface structures, while finer stages recover local details, object boundaries, and depth discontinuities.
This result explains the motivation for predicting hierarchical residual maps instead of decoding depth as a long raster-scan token sequence.


\begin{table}[!ht]
    \centering
    \caption{Effect of AR resolution. Higher AR resolution preserves finer generation stages and improves both metric and relative depth estimation.}
    \label{tab:ar_resolution}
    \begin{tabular}{c|cc|c}
    \toprule
    \multirow{2}{*}{AR Resolution} 
    & \multicolumn{2}{c|}{NYUv2} 
    & DA-2K \\
    \cline{2-4}
    & AbsRel$\downarrow$ & $\delta_1 \uparrow$ & Acc.$\uparrow$ \\
    \midrule
    512  & 5.3 & 96.7 & 89.6 \\
    1024 & \textbf{4.8} & \textbf{97.5} & \textbf{95.8} \\
    \bottomrule
    \end{tabular}
\end{table}

\paragraph{Effect of AR resolution}
We further analyze the impact of AR resolution by reducing the default generation resolution from 1024 to 512.
As shown in Tab.~\ref{tab:ar_resolution}, the lower-resolution variant degrades on both metric and relative depth estimation.
On NYUv2, AbsRel increases from 4.8 to 5.3, while $\delta_1$ decreases from 97.5 to 96.7.
On DA-2K, relative depth accuracy drops more significantly from 95.8\% to 89.6\%.
This suggests that fine generation stages are especially important for preserving local spatial relationships and pairwise depth ordering.
This result is consistent with the coarse-to-fine design of \methodName{}.
A higher AR resolution preserves more fine-grained generation stages, enabling the model to refine local structures after establishing the coarse layout.
When the resolution is reduced, the generation process terminates at a coarser spatial level, which weakens boundary recovery and local geometric refinement.

\begin{table}[!ht]
    \centering
    \caption{Model scaling and efficiency analysis. Runtime and memory are measured on a single NVIDIA A800 GPU with stage-wise decoding.}
    \label{tab:model_scaling}
    \resizebox{\columnwidth}{!}{
    \begin{tabular}{l|cc|cc|cc}
    \toprule
    \multirow{2}{*}{Model} 
    & \multicolumn{2}{c|}{NYUv2} 
    & \multicolumn{2}{c|}{KITTI}
    & \multirow{2}{*}{Time}
    & \multirow{2}{*}{Memory} \\
    \cline{2-5}
    & AbsRel$\downarrow$ & $\delta_1 \uparrow$ 
    & AbsRel$\downarrow$ & $\delta_1 \uparrow$
    & & \\
    \midrule
    Marigold~\cite{Marigold} & 5.5 & 96.4 & 9.9 & 91.6 & -- & -- \\
    \midrule
    ARDepth-2B & 5.1 & 96.9 & 8.9 & 92.6 & $\sim$0.7s & $\sim$8GB \\
    ARDepth-8B & \textbf{4.8} & \textbf{97.5} & \textbf{8.4} & \textbf{93.5} & $\sim$3.4s & $\sim$24GB \\
    \bottomrule
    \end{tabular}
    }
\end{table}

\subsection{Model Scaling and Efficiency Analysis}
\label{sec:exp-scaling}

To examine whether the gains of \methodName{} mainly come from model scale, we compare two variants, ARDepth-2B and ARDepth-8B, under the same hierarchical AR formulation and conditioning design.
As shown in Tab.~\ref{tab:model_scaling}, increasing the model size from 2B to 8B leads to consistent improvements on both NYUv2 and KITTI.
On NYUv2, AbsRel decreases from 5.1 to 4.8, and $\delta_1$ improves from 96.9 to 97.5.
On KITTI, AbsRel decreases from 8.9 to 8.4, and $\delta_1$ improves from 92.6 to 93.5.
These results show that larger capacity improves the representation ability of the hierarchical AR backbone.

More importantly, the 2B variant already outperforms Marigold on the reported benchmarks.
This indicates that the improvement is not solely attributable to the 8B backbone; even with a smaller parameter scale, the proposed hierarchical generation process and task-aware conditioning provide clear gains over a representative diffusion-based baseline.
The 8B model further improves performance, suggesting that \methodName{} benefits from model scaling while retaining architectural effectiveness at a smaller scale.

We also report inference cost on a single NVIDIA A800 GPU.
With stage-wise decoding and KV caching, ARDepth-2B requires approximately $\sim$0.7s per sample with about $\sim$8GB peak memory, while ARDepth-8B requires approximately $\sim$3.4s per sample with about $\sim$24GB peak memory.
Although the current implementation remains more computationally demanding than lightweight discriminative MDE models, the hierarchical formulation avoids decoding the full dense representation as a single long raster-scan sequence, making it more practical than a naive flat AR decoding design in terms of sequence structure.
Further acceleration through quantization, token pruning, speculative decoding, caching optimization, or parallel decoding is orthogonal to the proposed formulation and can be explored in future implementations.

\subsection{Discussion and Limitations}

Although \methodName{} demonstrates the potential of hierarchical AR generation for MDE, several limitations remain. First, the current 8B model still incurs non-negligible inference cost due to sequential AR decoding, although the 2B variant provides a lighter alternative. Second, SAG relies on off-the-shelf VLM descriptions, which may occasionally contain noisy spatial details. For this reason, \methodName{} uses compact scene-level descriptions rather than explicit textual depth reasoning. Finally, the current formulation focuses on single-image depth estimation. Extending scale-progressive AR generation to temporally consistent video depth or multi-view geometric reconstruction is an interesting future direction.

\section{Conclusion}
We revisit monocular depth estimation from a constructive perspective, exploring auto-regressive generation as a structured pathway for modeling geometric signals. Instead of treating depth as a globally smooth field recovered through implicit refinement, we model it as a hierarchically organized representation whose spatial dependencies emerge progressively across scales.
The core contribution of \methodName{} lies in introducing task-aware architectural components, namely, the scale-progressive visual conditioning and the semantic-aware guidance that explicitly factorize geometric reasoning while preserving global consistency.
Rather than positioning auto-regressive modeling as a replacement for diffusion-based approaches, our work demonstrates that, with task-aware architectural design, auto-regressive frameworks can also effectively capture dense geometric structures.
More broadly, this perspective suggests a potential direction for extending auto-regressive foundation models (e.g., MLLMs) with geometric perception capabilities, enabling unified modeling of spatial reasoning and multi-modal understanding.

\section*{Acknowledgments}
This work is supported in part by the National Key R\&D Program of China (NOs.2024YFB3908503 and 2024YFB3908500), in part by the National Natural Science Foundation of China under Grants 62322608, and in part by the PCL Major Key Project under Grant PCL2025A17-2. The authors would like to thank the National Supercomputer Center in Guangzhou for providing high-performance computing resources.



\bibliographystyle{IEEEtran}
\bibliography{refs}

\vfill

\end{document}